\documentclass[conference]{IEEEtran}

\usepackage{cite}
\usepackage{amsmath,amssymb,amsfonts}
\usepackage{amsthm}
\newtheorem{definition}{Definition}
\theoremstyle{definition}

\usepackage{algorithmic}
\usepackage{graphicx}
\usepackage{textcomp}
\usepackage{xcolor}
\usepackage{subcaption}
\usepackage{multirow}
\usepackage{booktabs}

\def\BibTeX{{\rm B\kern-.05em{\sc i\kern-.025em b}\kern-.08em
    T\kern-.1667em\lower.7ex\hbox{E}\kern-.125emX}}
\begin{document}

\title{Metric Learning Improves the Ability of Combinatorial Coverage Metrics to Anticipate Classification Error}

\author{\IEEEauthorblockN{Tyler Cody}
\IEEEauthorblockA{\textit{National Security Institute} \\
\textit{Virginia Tech}\\
Arlington, VA, USA \\
}
\and
\IEEEauthorblockN{Laura Freeman}
\IEEEauthorblockA{\textit{National Security Institute} \\
\textit{Virginia Tech}\\
Arlington, VA, USA \\
}
}

\maketitle

\begin{abstract}

Machine learning models are increasingly used in practice. However, many machine learning methods are sensitive to test or operational data that is dissimilar to training data. 
Out-of-distribution (OOD) data is known to increase the probability of error and research into metrics that identify what dissimilarities in data affect model performance is on-going. 
Recently, combinatorial coverage metrics have been explored in the literature as an alternative to distribution-based metrics. Results show that coverage metrics can correlate with classification error. 
However, other results show that the utility of coverage metrics is highly dataset-dependent. In this paper, we show that this dataset-dependence can be alleviated with metric learning, a machine learning technique for learning latent spaces where data from different classes is further apart. 
In a study of 6 open-source datasets, we find that metric learning increased the difference between set-difference coverage metrics (SDCCMs) calculated on correctly and incorrectly classified data, 
thereby demonstrating that metric learning improves the ability of SDCCMs to anticipate classification error.  
Paired t-tests validate the statistical significance of our findings. 
Overall, we conclude that metric learning improves the ability of coverage metrics to anticipate classifier error and identify when OOD data is likely to degrade model performance. 

\end{abstract}


\section{Introduction}

Machine learning is an increasingly dominant part of software systems and, thus, is of increasing interest in the software testing community. Compared to the standard statistical performance analysis on independent, identically-distributed data that machine learning practitioners and researchers typically use to test machine learning \cite{arlot2010survey}, the perspective of the software testing community has a broader scope. The learned models produced by machine learning algorithms are function approximations, and, as such, can be treated with the general test methods developed for system functions \cite{tripathy2011software}. A key set of general test methods identified for software functions is combinatorial interaction testing \cite{kuhn2013combinatorial}.

Combinatorial interaction testing is typically concerned with the interaction of various system components and typically applied during system integration to determine the subset of interactions among components that cause the majority of failures. It has been shown reported that most software faults can be attributed to 2-, 3-, or 4-way interactions among software components \cite{kuhn2004software}. Recently, combinatorial interaction testing and related concepts have been extended to machine learning. A key concept in related machine learning literature is combinatorial coverage \cite{kuhn2013combinatorial}.

Combinatorial coverage measures the number of interactions present in a dataset against the total number of possible interactions. Early applications of combinatorial coverage to machine learning include the use of coverage within deep learning algorithms \cite{pei2017deepxplore, ma2018deepgauge, ma2019deepct}. These early works had mixed results, and faced heavy criticism \cite{li2019structural}. Central to critiques was the idea that combinatorial coverage was concerned with the total number of possible interactions, or, otherwise put, the entire universe of interactions. Most of these interactions are not natural, and many can be even be considered adversarial \cite{li2019structural}.

While coverage metrics faced difficulties within machine learning algorithms, their application to the testing of machine learning algorithms grew in the literature. Early examples include use cases in explainability \cite{kuhn2020combinatorial}, autonomous vehicles \cite{tuncali2018simulation}, and test set construction \cite{ackerman2020freaai}. Building on this early work, Lanus et. al proposed set-difference combinatorial coverage metrics (SDCCMs) which, in contrast to traditional coverage metrics \cite{kuhn2013combinatorial}, did not calculate coverage with respect to the entire universe of interactions, but rather with respect to another dataset \cite{lanus2021combinatorial}. SDCCMs can be used to compare the interactions seen in training, to those seen in testing, to those seen in operation---all relative to each other, not relative to all possible interactions. Cody et. al use SDCCMs to directly refute previous critiques of the use of combinatorial coverage in machine learning using MNIST image data \cite{cody2022systematic}.

While previous work has affirmed the utility of combinatorial coverage concepts to machine learning, their use has also been shown to be highly dataset-dependent \cite{kuhn2022combination}. In this work, we propose that metric learning be used to address dataset-dependence. Metric learning is a subfield of machine learning concerned with learning latent spaces where data from different classes is further apart than in their original space. In a study of 6 open-source datasets from a variety of domains and with a variety of sizes, we provide evidence that metric learning improves the ability of SDCCMs to anticipate classification error. In doing so, we also provide supporting evidence for existing claims in the literature that SDCCMs are correlated to classification error \cite{lanus2021combinatorial, cody2022systematic}.

This paper is structured as follows. First, related work is reviewed. Combinatorial coverage and set-difference combinatorial coverage (SDCC) are defined as given by \cite{lanus2021combinatorial}. Then, background is provided on metric learning and details are provided for the metric learning methods used in this paper. The method of analysis is described, results are presented with accompanying discussion, and, lastly, conclusions are drawn.







\section{Preliminaries}
\label{sec:related}

Notions of combinatorial coverage have been applied to deep learning \cite{pei2017deepxplore, ma2018deepgauge, ma2019deepct}. These applications have been critiqued \cite{li2019structural}. Broader use of coverage concepts in machine learning include explainability \cite{kuhn2020combinatorial, chandrasekaran2021combinatorial}, autonomous vehicles \cite{tuncali2018simulation, chandrasekaran2021combinatorialv}, test set construction \cite{ackerman2020freaai}, fairness testing \cite{patel2022combinatorial}, physical unclonable functions \cite{kuhn2020combinatorial}, multi-domain operations \cite{cody2022combinatorial}, and active learning \cite{sai2022active}. Many works focus on traditional measures of coverage \cite{kuhn2013combinatorial}, however recent work proposes new measures for applications of coverage to machine learning based on set-difference \cite{lanus2021combinatorial}.

\subsection{Coverage Formalism}

In combinatorial interaction testing for machine learning, \emph{factors} are the features $\mathcal{X}$ and classes $\mathcal{Y}$ and the \emph{values} of those factors are their events. Note, in combinatorial interaction testing, continuous-valued factors must be discretized to a finite set of values. A $t$-way \emph{value combination} is a $t$-tuple of (factor, value) pairs. If there are $k$ factors in $\mathcal{X} \times \mathcal{Y}$, then each element $(x, y) \in \mathcal{X} \times \mathcal{Y}$ contains $k \choose t$ $t$-way value combinations. Value combinations are a formal description of interactions.

\emph{Combinatorial coverage}, sometimes termed total $t$-way coverage, measures the proportion of valid $t$-way value combinations that appear in a set \cite{kuhn2013combinatorial}. Value combinations that appear are considered \emph{covered}. Combinatorial coverage is defined in the following.

\begin{definition}{\emph{$t$-way Combinatorial Coverage.}} \\
    Consider a universe with $k$ factors such that $\mathcal{U}$ is the set of all valid $k$-way value combinations. Let $\mathcal{U}^t$ be the set of valid $t$-way combinations. Given a set of data $D \subseteq \mathcal{U}$, let $D^t$ define the set of $t$-way value combinations appearing in $D$. The $t$-way combinatorial coverage of $D$ is
    $$CC^t(D) = \frac{|D^t|}{|\mathcal{U}^t|},$$
    where $|D|$ denotes the cardinality of $D$. 
\end{definition}

SDCC is alternatively measures the proportion of valid $t$-way value combinations that appear in one set relative to another set. SDCC is defined in the following \cite{lanus2021combinatorial}.

\begin{definition}{\emph{$t$-way Set Difference Combinatorial Coverage.}} \\
    Let $D_A$ and $D_B$ be sets of data, and $D{_A}^t$ and $D{_B}^t$ be the corresponding $t$-way sets of data. The set difference $D{_B}^t \setminus D{_A}^t$ gives the value combinations that are in $D{_B}^t$ but that are not in $D{_S}^t$. The $t$-way set difference combinatorial coverage is
    $$SDCC^t(D_B, D_A) = \frac{|D{_B}^t \setminus D{_A}^t|}{|D{_B}^t|}.$$
\end{definition}
\noindent $SDCC^t$ is bounded $[0, 1]$ where $1$ indicates no overlap, i.e., $D{_B}^t \cap D{_A}^t = \emptyset$, and $0$ indicates that $D{_B}^t \subseteq D{_A}^t$. In other words, if $D_B$ is the testing data and $D_A$ is the training data, $0$ indicates that all testing combinations are present in the training data and $1$ indicates that none are present. Note, we use the term SDCCMs to refer to SDCC over a set of t-values.

\section{Metric Learning Background}

Metric learning is generally concerned with learning distance functions \cite{kulis2013metric}. This typically involves learning a transformation to a latent space where similar data are closer together and dissimilar data are farther apart. As such, the latent spaces learned by metric learning induce sparsity in the input space that is both useful for making predictions (by making data more linearly separable). This may also be useful for coverage as the more separated data is easier to discretize \cite{ramirez2016data}. Although deep metric learning methods are an active area of interest \cite{kaya2019deep, roth2020revisiting}, critiques suggest that, so far, deep metric learning offers little improvement over classical methods in cases where both deep and classical methods can be applied \cite{musgrave2020metric}.

We use three of the most prevalent classical methods: Neighborhood Component Analysis (NCA) \cite{goldberger2004neighbourhood}, Metric Learning for Kernel Regression (MLKR) \cite{weinberger2007metric}, and Large Margin Nearest Neighbor (LMNN) \cite{weinberger2009distance}.
\begin{itemize}
    \item NCA is a variation on k-nearest neighbors (KNN) classification that directly maximizes a variant of leave-one-out performance.
    \item MLKR learns a distance function by directly minimizing the leave-one-out regression error.
    \item LMNN is a variation on KNN classification that learns a Mahalanobis distance.
\end{itemize}
These three methods are supervised metric learning methods because they use class labels during training.

\section{Data}

In this paper, 6 open-source datasets are used. All datasets can be retrieved from the University of California Irvine (UCI) Machine Learning Repository \cite{asuncion2007uci}. The datasets are described in Table \ref{table:datasets} and are: the ``Wine Data Set'' (Wine), the ``Rice (Cammeo and Osmancik) Data Set'' (Rice), the ``Yeast Data Set'' (Yeast), the ``Car Evaluation Data Set'' (Car), the ``Breast Cancer Wisconsin (Diagnostic) Data Set'' (Cancer), and the ``Balance Scale Data Set'' (Balance). The datasets vary in size, number of classes, number of features, and feature type. The Cancer dataset is reduced from its original feature space to only the `mean' features. All other datasets are unaltered.

\begin{table}[ht]
\captionsetup{justification=centering, labelsep=newline, font=footnotesize}
\centering
\caption{\sc Dataset Information}
\begin{tabular}{l c c c c} 
 \toprule
 Dataset & Sample Size & Classes & Features & Feature Type \\ 
 \cmidrule(lr){1-1}
 \cmidrule(lr){2-2}
 \cmidrule(lr){3-3}
 \cmidrule(lr){4-4}
 \cmidrule(ll){5-5}
 Wine & 178 & 3 & 13 & Continuous \\ 
 Rice & 3810 & 2 & 7 & Continuous \\
 Yeast & 1484 & 10 & 8 & Continuous \\
 Car & 1728 & 4 & 6 & Discrete \\
 Cancer & 569 & 2 & 10 & Continuous \\
 Balance & 625 & 3 & 4 & Discrete \\
 \bottomrule
\end{tabular}
\label{table:datasets}
\end{table}

\section{Method}

In this paper, we investigate whether metric learning improves the ability of SDCCMs to anticipate classification error. To assess the ability to anticipate error, we measure the difference between SDCCMs calculated (1) between training data and correctly classified data and (2) between training data and incorrectly classified data. We suppose that the larger the difference, the better SDCCMs perform as a metrics for anticipating classification error. To conduct our analysis, the data must be prepared.

The data is prepared for coverage analysis as follows. The NCA, MLKR, and LMNN algorithms are used to learn transformations to latent spaces using all available data. The metric learning methods use the default settings of metric-learn\cite{metric-learn}. Each dimension of the latent spaces is used as a feature. The number of dimensions is equal to the number of features in the original data. Then, the NCA, MLKR, LMNN, and original spaces are discretized. Note, the Car and Balance datasets have discrete features and thus the discretization is not applied to their original spaces. The discretization process treats each feature separately. Each feature's data is clustered using the standard k-means clustering algorithm from scikit-learn with k=5 bins \cite{pedregosa2011scikit}.

After transformation and discretization, the data is ready for coverage analysis. The coverage analysis is conducted as follows. For each dataset, 10 folds are created with 80\%-20\% train-test splits using random sampling. For each fold, a decision tree (DT), support vector machine (SVM), and KNN classifier is trained and tested. The classifiers use the default scikit-learn settings \cite{pedregosa2011scikit}, except the maximum depth is set to 2 for the DT classifier. The correctly and incorrectly classified test data are separated. Then, the SDCCMs are calculated over the discretized NCA, MLKR, LMNN, and original features (i.e., not the labels) (1) between the training data and the correctly classified data and (2) between the training data and the incorrectly classified data. In both cases, the SDCC is used to measure the ratio of the number of interactions that appear in the training data that do not appear in test data to the total number of interactions that appear in the training data. In previous literature \cite{cody2022systematic} and during initial testing, this direction of set-difference showed stronger correlation with error than the ratio of interactions appearing in the test data that do not appear in the training data to the total number of interactions in the test data. The difference between SDCCMs calculated in (1) and (2) is assessed for $t = (2, 3, 4)$.

\section{Results}

\begin{figure*}[htb]
    \centering
    \includegraphics[width=.95\textwidth]{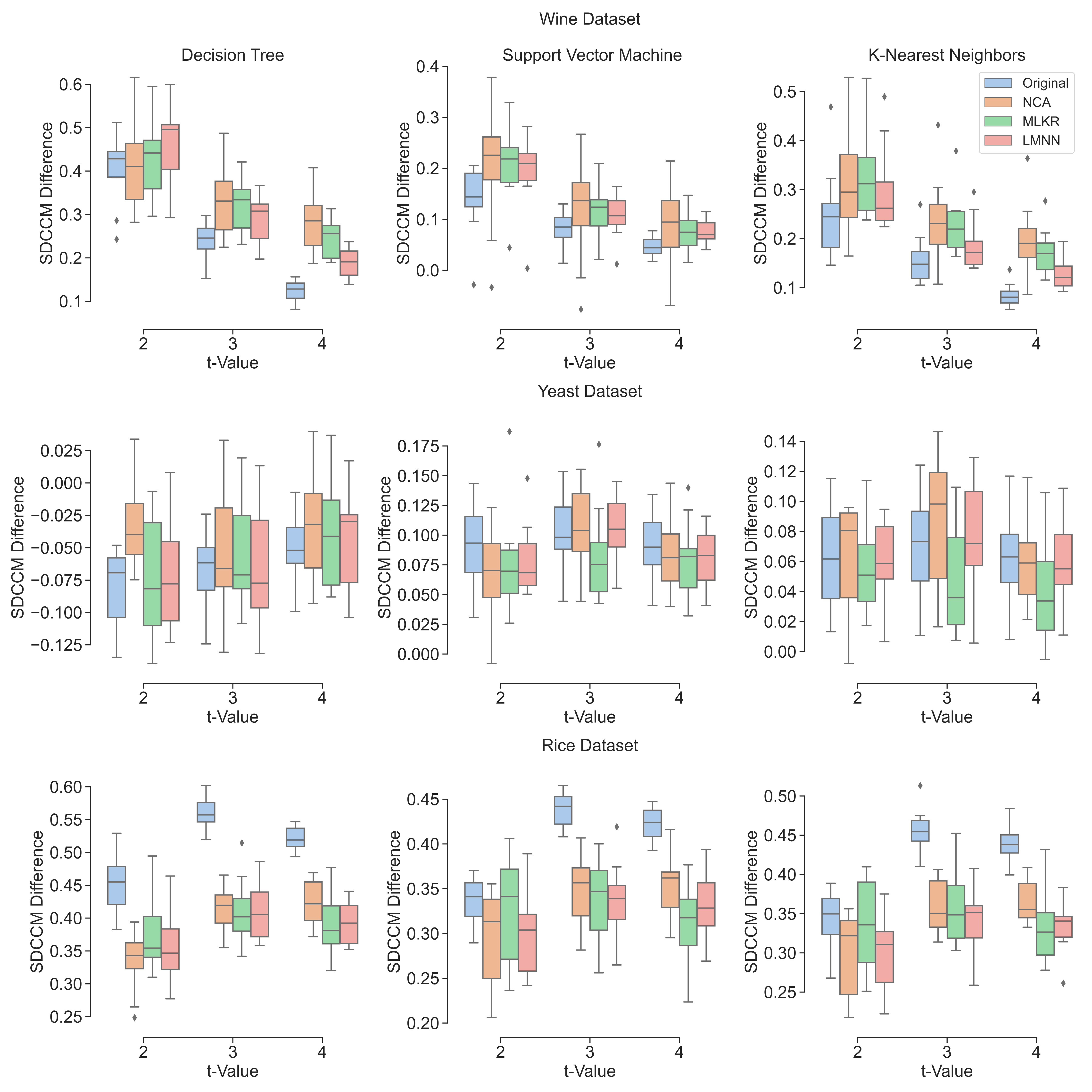}
    \caption{Results by dataset and classifier for Wine, Yeast, and Rice.}
    \label{fig:wine_boxplot}
\end{figure*}

\begin{figure*}[htb]
    \centering
    \includegraphics[width=.95\textwidth]{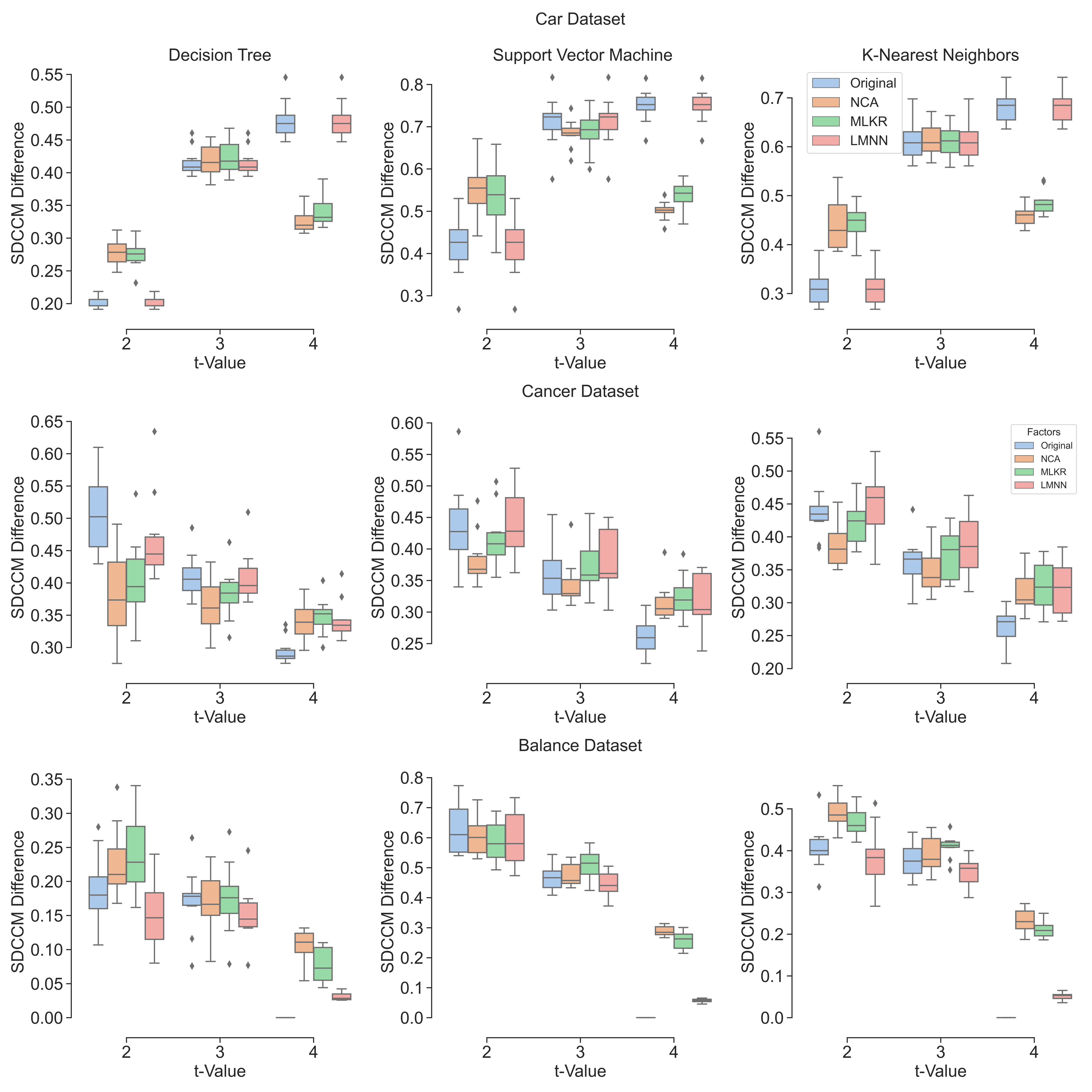}
    \caption{Results by dataset and classifier for Car, Cancer, and Balance.}
    \label{fig:car_boxplot}
\end{figure*}

The results of the coverage analysis are examined visually and statistically to assess whether metric learning increases the ability of SDCCMs to anticipate classification error. The accuracy scores of the classifiers are summarized in Table \ref{table:performances}. Figures \ref{fig:wine_boxplot} and \ref{fig:car_boxplot} visualize the results. The rows of the figures correspond to the datasets. The columns of the figures correspond to the DT, SVM, and KNN classifiers, respectively, from left-to-right. 

\begin{table}[t]
\captionsetup{justification=centering, labelsep=newline, font=footnotesize}
\centering
\caption{\sc Classification Accuracy}
\begin{tabular}{l c c c c} 
 \toprule
 n = 10 & DT & SVM & KNN & Random \\ [0.5ex]
 Dataset & (Mean, Std) & (Mean, std) & (Mean, Std) & 1/No. Classes \\ 
 \cmidrule(lr){1-1}
 \cmidrule(lr){2-2}
 \cmidrule(lr){3-3}
 \cmidrule(lr){4-4}
 \cmidrule(ll){5-5}
 Wine & (0.85, 0.07) & (0.61, 0.06) & (0.72, 0.08) & 0.33 \\
 Rice & (0.92, 0.01) & (0.88, 0.14) & (0.88, 0.02) & 0.50 \\
 Yeast & (0.48, 0.04) & (0.59, 0.03) & (0.56, 0.03) & 0.10 \\
 Car & (0.78, 0.02) & (0.93, 0.02) & (0.90, 0.02) & 0.25 \\
 Cancer & (0.92, 0.02) & (0.89, 0.03) & (0.89, 0.03) & 0.50 \\
 Balance & (0.64, 0.04) & (0.90, 0.03) & (0.82, 0.03) & 0.33 \\
 \bottomrule
\end{tabular}
\label{table:performances}
\end{table}

Within each subplot, the y-axis corresponds to the SDCC for the incorrectly classified data subtracted from the SDCC for the correctly classified test data. Higher values along the y-axis indicate a higher ability to anticipate classification error. The x-axis corresponds to the t-values used in the SDCC calculations, and boxplots are grouped according to the t-values. The different colors of the boxplots correspond to the discretized original, NCA, MLKR, and LMNN spaces, respectively, from left-to-right. The center line of the boxes indicates the median, the upper and lower borders of the boxes indicate the upper and lower quartiles, and the whiskers indicate the rest of the distribution. Outliers are depicted as black diamonds. Note, these outliers are identified by their distance from the inter-quartile range, and are not removed in later statistical analysis.

The general trends in the figures can be described as follows. In Figure \ref{fig:wine_boxplot}, the Wine dataset shows outperformance by metric learning for nearly all classifiers, t-values, and metric learning methods. The Yeast dataset, in contrast, depicts mixed results, with at least one metric learning method outperforming in many cases, but no cases of all metric learning methods outperforming. Lastly, the Rice dataset shows the opposite results of the Wine dataset, with the original space outperforming for nearly all cases. This dataset-to-dataset variance in results is expected, and has been noted in previous literature \cite{kuhn2022combination}. The trends in Figure \ref{fig:car_boxplot} are more nuanced. For the Car dataset, metric learning spaces tend to outperform the original space more for lower values of t, while for the Cancer dataset, metric learning outperforms for higher values of t. The Balance dataset shows both of these trends, with, in general, more outperformance occurring for $t=2$ and $t=4$ than for $t=3$.

In summary, the figures show varied results, however, for many cases, metric learning clearly increases the ability of SDCCMs to anticipate classification error. From visual inspection of Figures \ref{fig:wine_boxplot} and \ref{fig:car_boxplot}, the largest dependency in the success of metric learning appears to be on the dataset, followed by smaller dependencies on t-values and classifiers, respectively. It should be noted, that across all datasets, classifiers, and t-values, only the Yeast dataset with the DT classifier shows near-zero difference between the SDCCMs for correctly and incorrectly classified test data. This provides evidence for claims in previous literature that SDCCMs can be used to anticipate error in machine learning \cite{lanus2021combinatorial}. While Cody et. al \cite{cody2022systematic} demonstrated this correlation on MNIST image data, herein, we evidence the presence of a more general trend by using data from 6 different domains.

To better assess whether metric learning increases the ability of SDCCMs to anticipate classification error, we conduct a rigorous statistical analysis using the paired t-test. The paired t-test assesses whether or not the pairwise differences between two sets of values varies from $0$ \cite{hsu2014paired}. If the test statistic is greater or less than $0$ and the p-value is deemed significant, than the null hypothesis that there is no difference is deemed false. For our analysis, we treat test results with a p-value $< 0.05$ as significant. We pairwise compare the SDCCM differences of each metric learning space with those of the original space. We apply the paired t-test for each dataset and t-value since these are the primary dependencies identified in the boxplots.

In Table \ref{table:statistics}, the results of the statistical analysis are shown with two significant figures. Since our hypothesis is that metric learning improves the ability of SDCCMs to anticipate classification error, as opposed to a specific metric learning method, the table identifies the metric learning method with the highest test statistic for each dataset and t-value. For the Car dataset at t-values of 2 and 3, the LMNN and original spaces were identical, and thus there was no pairwise difference. The results of Table \ref{table:statistics} are summarized in Table \ref{table:summary}. In $15/18$ or 83\% of cases, there was a metric learning space that was as good as or better than the original space. In $10/18$ or 56\% of cases, there was a metric learning space that was better than the original space. In conclusion, these results provide evidence that metric learning improves the ability of SDCCMs to anticipate classification error.

\begin{table*}[h]
\captionsetup{justification=centering, labelsep=newline, font=footnotesize}
\centering
\caption{\sc Paired T-Tests for Each Dataset and t-Value}
\begin{tabular}{l c c c c c c c c c c c c} 
 \toprule
 n = 30 & \multicolumn{4}{c }{t = 2} & \multicolumn{4}{c }{t = 3} & \multicolumn{4}{c }{t = 4} \\ 
 \cmidrule(lr){2-5}
 \cmidrule(lr){6-9}
 \cmidrule(ll){10-13}
 Dataset & Statistic & p-Value & Result & Method & Statistic & p-Value & Result & Method & Statistic & p-Value & Result & Method \\ 
 \cmidrule(lr){1-1}
 \cmidrule(lr){2-5}
 \cmidrule(lr){6-9}
 \cmidrule(ll){10-13}
 Wine & 8.2 & 5.2e-9 & Higher & LMNN & 9.0 & 6.0e-10 & Higher & MLKR & 13 & 2.6e-13 & Higher & LMNN \\
 Rice & -2.5 & 2.1e-2 & Lower & MLKR & -12 & 1.8e-12 & Lower & MLKR & -13 & 2.7e-13 & Lower & NCA \\
 Yeast & 1.1 & 2.9e-1 & Same & NCA & 2.2 & 3.5e-2 & Higher & NCA & 0.70 & 5.0e-1 & Same & NCA \\
 Car & 18 & 1.2e-17 & Higher & NCA & 0 & 0 & Same & LMNN & 0 & 0 & Same & LMNN \\
 Cancer & -0.9 & 3.8e-1 & Same & LMNN & 2.3 & 2.9e-2 & Higher & LMNN & 13 & 2.5e-13 & Higher & MLKR \\
 Balance & 2.7 & 1.1e-2 & Higher & NCA & 4.8 & 4.0e-5 & Higher & MLKR & 20 & 3.5e-18 & Higher & LMNN \\
 \bottomrule
\end{tabular}
\label{table:statistics}
\end{table*}

\begin{table}[h]
\captionsetup{justification=centering, labelsep=newline, font=footnotesize}
\centering
\caption{\sc Paired T-Test Summary}
\begin{tabular}{l c c c | c} 
 \toprule
 Dataset & t = 2 & t = 3 & t = 4 & Totals\\ 
 \cmidrule(lr){1-1}
 \cmidrule(lr){2-2}
 \cmidrule(lr){3-3}
 \cmidrule(lr){4-4}
 \cmidrule(ll){5-5}
 Wine & Higher & Higher & Higher & \textit{Higher} \\ 
 Rice & Lower & Lower & Lower & 10 \\
 Yeast & Same & Higher & Same & \textit{Same} \\
 Car & Higher & Same & Same & 5 \\
 Cancer & Same & Higher & Higher & \textit{Lower} \\
 Balance & Higher & Higher & Higher & 3 \\
 \bottomrule
\end{tabular}
\label{table:summary}
\end{table}

\section{Discussion}

\begin{figure*}[htb]
    \centering
    \includegraphics[width=.75\textwidth]{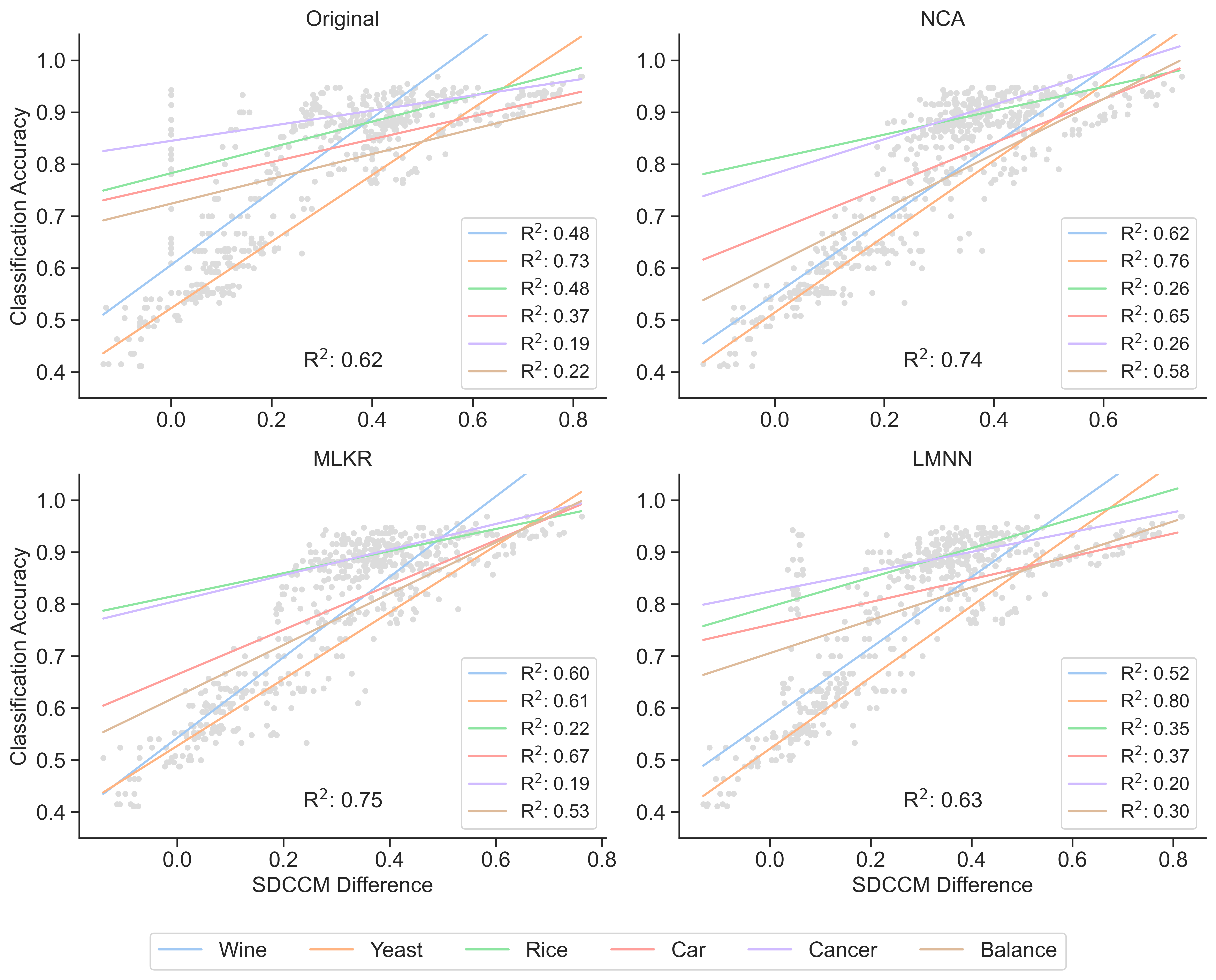}
    \caption{Correlation analysis of classification accuracy and SDCCM differences.}
    \label{fig:corr_plot}
\end{figure*}

Results are discussed in detail in the previous section, however, the role of classification accuracy was not considered. Figure \ref{fig:corr_plot} shows the classification accuracy plotted against the SDCCM differences for the metric learning and original spaces. The points are from all datasets, t-values, and classifiers. The R$^2$ correlation between the accuracies and differences for all data (for each respective space) is displayed in the subplots. Also, regression lines are calculated for each dataset using linear least-squares regression. The R$^2$ correlations given by these regressions are displayed in the legend boxes of each subplot. The regression lines and their correlations are labeled by their dataset in the legend at the bottom of the figure.

Immediately, we can note that there are clear positive correlations between accuracy and SDCCM differences in all spaces. This means that the ability of SDCCM to anticipate classifier error increases with classification accuracy. Also, it appears that NCA and MLKR are able to transform points away from the upper-left quadrant, leading to higher correlations. These upper-left points indicate cases where SDCCM differences are uncorrelated to classification accuracy. This means that the SDCC calculations for correctly and incorrectly classified test data are similar regardless of classification accuracy.

It may seem undesirable to have points in the upper-left quadrant. But this intuition requires nuance. Consider the distinct goals of learning classifiers and learning latent spaces for coverage analysis. First, note that in this paper we are studying whether metric learning provides more meaningful spaces over which to conduct coverage analysis. Now, setting that motivation aside for a moment, consider that one has obtained a meaningful space---a space where no available method offers significant improvement in the ability to anticipate errors using SDCCMs. In such a space, the presence of points in the upper-left quadrant that indicate high performance but a low difference between SDCCMs calculated on correctly and incorrectly classified test data can be taken as a signal that classification errors are occurring at random, or at least without significant bias towards any particular region in the space of interactions. In contrast, the presence of points in the upper-right quadrant indicate that there is such a bias; that is, that even though classification accuracy is high, there are still regions in the space of interactions where the classifier is predictably wrong.

Therefore, NCA and MLKR, by finding transformations which shift points away from the upper-left quadrant, are effectively finding new representations for the space of interactions to which the classifiers are predictably wrong or otherwise biased. From this reasoning, it seems that the goal of metric learning for coverage analysis should be to increase the correlation between model performance and SDCCM differences, while the goal of learning classifiers should be to decrease the correlation by shifting test points to the upper-left quadrant.

There are a few interesting trends in the dataset-specific correlations. Rice and Cancer, the two datasets where metric learning performed the worst based on visual inspection of Figures \ref{fig:wine_boxplot} and \ref{fig:car_boxplot}, consistently have the lowest correlations. Moreover, for Rice, the dataset where metric learning performed the worst based on the statistical analysis summarized in Table \ref{table:summary}, the original space has a significantly higher correlation than any metric learning space. And for Cancer, the correlations are similar between the original and metric learning spaces. In contrast, Wine and Balance---the best performing metric learning methods based on Table \ref{table:summary}---have correlations that are higher for the metric learning spaces than for the original space. From these observations, it seems that the larger the difference there is in like-for-like correlations between original and metric learning spaces, the more metric learning improves the ability of SDCCMs to anticipate error. This trend, however, needs more datasets for further investigation.

\section{Conclusion}

In this paper, we provide evidence that metric learning can improve the ability of SDCCMs to anticipate classification error. Importantly, the metric learning spaces were minimally tuned. This suggests that the demonstrated outperformance by metric learning is not over-engineered and can likely increase further with hyperparameter optimization. This promising evidence supports arguments for combinatorial coverage in identifying when out-of-distribution (OOD) data is likely to degrade model performance.

Additionally, our experiments support previous claims in the literature the SDCCMs correlate to the error of machine learning classifiers. Despite past criticisms of particular applications of coverage concepts to neural networks, there is a growing body of literature supporting the use of combinatorial coverage concept in machine learning in general. As this marked trend continues, it is important to revisit the use of coverage concepts in learning algorithm development, not just auxiliary support systems like explainability, fairness, or error anticipation.

Metric learning offers a promising opportunity in this direction. Metric learning algorithms should be revisited in future work for their prospective use in coverage analysis. A particular concern in this area is metric learning methods that inherently learn discretized or readily discretizable spaces. Further, additional terms can be considered in the loss functions of metric learning algorithms such that the learned spaces are well-suited for particular kinds of coverage analysis, as opposed to, e.g., the traditional focus on being well-suited for classification and prediction problems. While the use of classical metric learning algorithms herein demonstrates an improvement in the ability to anticipate classification error, the results also indicate that significant dataset-dependence remains. Metric learning algorithms more tailored to coverage concepts could help further alleviate dataset-dependence in the utility of coverage analysis. 

\section*{Acknowledgements}

This material is based upon work supported, in whole or in part, by the U.S. Department of Defense through the Office of the Assistant Secretary of Defense for Research and Engineering (ASD(R\&E)) under Contract HQ003419D0003. The Systems Engineering Research Center (SERC) is a federally funded University Affiliated Research Center managed by Stevens Institute of Technology. Any views, opinions, findings and conclusions or recommendations expressed in this material are those of the author(s) and do not necessarily reflect the views of the United States Department of Defense nor ASD(R\&E).



\bibliographystyle{IEEEtran}
\bibliography{ref}

\end{document}